\documentclass[10pt,twocolumn,letterpaper]{article}

\usepackage{icb}
\usepackage{times}
\usepackage{booktabs}
\usepackage{lineno}
\usepackage[bookmarks=false]{hyperref}
\usepackage{amsmath,amssymb} 
\usepackage{diagbox}
\usepackage{color}
\usepackage{adjustbox}
\usepackage{diagbox}
\usepackage{siunitx}
\usepackage{epsfig}
\usepackage{subfig}
\usepackage{graphicx}
\usepackage{bm}
\usepackage{siunitx}
\usepackage{mathbbol} 
\usepackage[linesnumbered,ruled,vlined]{algorithm2e}
\usepackage[norule,symbol,perpage]{footmisc}

\newcommand\numberthis{\addtocounter{equation}{1}\tag{\theequation}}

\icbfinaltrue

\ificbfinal\fi

\newcommand{\CommaPunct}{\mathpunct{\raisebox{1.8ex}{,}}}

\begin{document}

\title{Occlusion-guided compact template learning for ensemble \\ deep network-based pose-invariant face recognition}

\author{Yuhang Wu and Ioannis A. Kakadiaris\\
Computational Biomedicine Lab, University of Houston\\
{\tt\small \{ywu35,ikakadia\}@central.uh.edu }
}

\maketitle
\thispagestyle{empty}

\begin{abstract}
	Concatenation of the deep network representations extracted from different facial patches helps to improve face recognition performance. However, the concatenated facial template increases in size and contains redundant information. Previous solutions aim to reduce the dimensionality of the facial template without considering the occlusion pattern of the facial patches. In this paper, we propose an occlusion-guided compact template learning (OGCTL) approach that only uses the information from visible patches to construct the compact template. The compact face representation is not sensitive to the number of patches that are used to construct the facial template, and is more suitable for incorporating the information from different view angles for image-set based face recognition. Instead of using occlusion masks in face matching (\eg, DPRFS \cite{Xu_2017_17643}), the proposed method uses occlusion masks in template construction and achieves significantly better image-set based face verification performance on a challenging database with a template size that is an order-of-magnitude smaller than DPRFS.
\end{abstract}

	\vspace{-0.5cm}
\section{Introduction}
	The problem of template learning in face recognition is to map a single facial image into a vector representation so that the distance between these representations reflects the actual similarity between facial images (or videos). In recent work, researchers have focused more on improving the discriminative capability of the template, and paid relatively less attention to the size of the template. The size of the template becomes important for large-scale face retrieval when the gallery database contains millions of images \cite{Wang_2017_17758}. A real-world surveillance scenario becomes infeasible to compute the similarity scores between one probe and all images in the gallery when it takes minutes to process a single video frame. Another application scenario is biometric credit cards, where the size of the template is limited by the memory of the chip inside it. In this work, we focus on learning a compact floating point template whose vector length should be as small as possible while preserving its discriminative capability.

\begin{figure}[!t]
    \includegraphics[width=7.8cm, height=3.5cm]{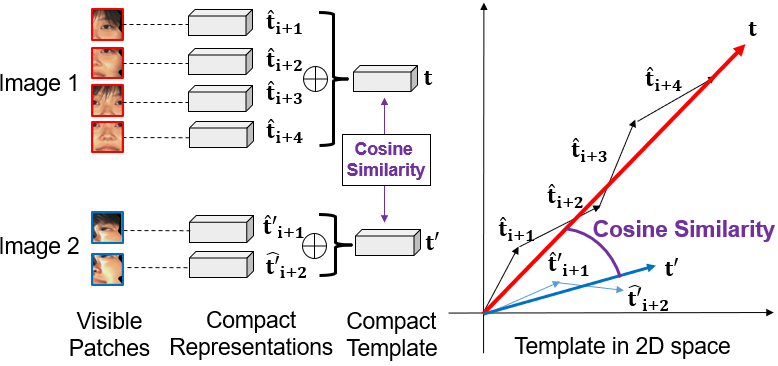} 
	\caption{Matching two compact templates generated by OGCTL. (L) The compact templates can be generated from different number of visible patches of two facial images. Compact facial representations are first extracted from visible patches. Then, they are added together with element-wise summation operation to create a compact but discriminative template. The compact facial templates from two images are compared based on the cosine similarity. (R) The compact templates are visualized in 2D space. The angular distance between the two templates determines the cosine similarity rather than the magnitude of the templates.
		\vspace{-0.6cm}
	}
	\label{DPRFS}

\end{figure}

	Learning a compact but highly discriminative template is challenging; the recent research Gong \etal \cite{IntrinsicDimension} indicates that template length directly impacts the discriminative capability of the template. To gather as much discriminative information as possible in the template, our starting point is based on an ensemble deep neural network model named DPRFS presented by \mbox{Xiang \etal \cite{Xu_2017_17643}}. Like other ensemble deep neural network models \cite{Sun_2014_16640, Sun_2014_17724, Sun_2015_16639, Hu_2015_17447, Liu_2015_16644, TPN},  DPRFS concatenates facial representations from multiple facial patches, but uses an occlusion vector to indicate which facial patch was occluded. It was demonstrated that the template generated by DPRFS achieves state-of-the-art results for single image based face matching under large head pose variation \cite{Xu_2017_17643}. However, it has the following limitations: (i) the template size is large (16K bytes), (ii) face matching speed is limited by selecting non-occluded face representations for similarity computations, (iii) it is hard to generate a single template that takes into account all the occlusion patterns of images in an image set. 
	
	To address the aforementioned limitations of the ensemble model, we invented an occlusion-guided compact template learning method to encourage the information integration between the face representations that are generated by each individual deep network. This operation preserves the discriminative co-occurrence pattern between multiple facial regions and significantly improves the performance of a compact template. The main contributions of this paper are: (i) we invented an occlusion-guided neural network architecture to learn a compact facial template from the face representations and occlusion masks generated by an ensemble model, (ii) we invented a method to generate a compact template from different number of visible facial patches. 
	
	\section{Related work}
In the literature on facial template learning, researchers have paid more attention to the performance of the templates rather than the template size. To the best of our knowledge, the size of the smallest template that achieves more than 99.5\% verification rate on LFW database \cite{Huang_2008_13354} is 0.5 \textit{KB} \cite{Liu_2015_16644, Schroff_2015_16462} (The data type of the template is assumed to be 32-bit single precision floating point as used in standard convolutional neural networks (CNN) \cite{FloatingPoint}). Although most of the selected algorithms reported near saturated verification performance on LFW database, their performances degrade in the databases that contain large head pose variations (e.g., UHDB31 database \cite{Le_2017_17704}).

Many of the recent face recognition systems employ ensemble models  \cite{Sun_2014_16640, Sun_2014_17724, Sun_2015_16639, Hu_2015_17447, Liu_2015_16644, TPN, Chellapa}, where face representations of multiple convolutional neural networks are concatenated together as the final facial template to enrich the template's representation capability. Hu \etal \cite{Hu_2015_17447} used extensive experiments to demonsrate that the concatenated representation consistently outperforms the output of a stand-alone CNN model. To learn diverse facial representation for concatenation, a common solution is to divide the facial image into multiple patches \cite{Sun_2014_16640, Sun_2014_17724, Sun_2015_16639, Hu_2015_17447, Liu_2015_16644, TPN}, and for each patch learn a separate CNN. However, these methods do not take into account the occlusion of patches under large head-pose variations. Some of the patches should be de-activated on-line due to facial occlusion because their corresponding features in the facial template are no longer informative. Using these non-discriminative face representations in face matching is detrimental to the final face recognition performance.

A compact template can be further compressed by principal component analysis (PCA) \cite{Jolliffe_1986_10826}, product quantization \cite{PQ}, or hashing-based binarization approaches \cite{FaceVideoHashing,DDH}. PCA is a well-known dimensionality reduction approach which has been applied in \cite{Sun_2014_16640, Sun_2014_17724} to reduce the facial template size from 76.8 \textit{KB} to 0.625 \textit{KB}.  Wang \cite{Wang_2017_17758} employs Product Quantization \cite{PQ} to convert a 1.25 \textit{KB} into a binary 64 Bytes template for large-scale face retrieval. Recently, deep hashing-based approaches were used to generate a binary template from the output of CNN with hashing layers \cite{FaceVideoHashing,DDH}. The common point of all these approaches is that they require a floating point face representation as the starting point to derive a compact facial template. The contribution of our approach is to provide a method to generate a compact face representation (0.5 \textit{KB} to 1 \textit{KB}) as a good candidate for the aforementioned studies \cite{Jolliffe_1986_10826,PQ,FaceVideoHashing,DDH}. This template is robust to facial self-occlusion caused by large head pose variations.

	\section{Method}

	\subsection{Template and occlusion mask}
DPRFS \cite{Xu_2017_17643} is designed to cope with the large head pose variation problem in face recognition. The algorithm is depicted in Fig. \ref{DPRFS}.  Before computing the face representations, a frontalized facial texture $\textbf{V}$ needs to be generated from the input image to normalize the pose variations \cite{Kakadiaris_2017_17717,Xu_2017_17643}. Let $\textbf{I}$ be an input image. A personalized facial model with the name AFM proposed by Kakadiaris \etal \cite{Kakadiaris_2007_5732} is initially estimated based on the method proposed by Dou \etal \cite{Dou_2017_17421}. The model is then fit onto $\textbf{I}$ based on the landmarks detected by Wu \etal \cite{Wu_2017_17445}. Then, a frontalized facial texture $\textbf{V}$ (UV image) is generated through the texture lifting process proposed by Toderici \etal \cite{Toderici_2010_9723}. Along with $\textbf{V}$, a binary self-occlusion map $\textbf{O}$ is generated based on the z-buffer as in \cite{Dou_2015_16379}, where each pixel in $\textbf{O}$ indicates whether the corresponding pixel of $\textbf{V}$ in $\textbf{I}$ is occluded or not. In DPRFS, $\textbf{V}$ is divided into $n$ semi-overlapped patches as depicted in \mbox{Fig. \ref{DPRFS}}. Let the set \mbox{$\mathbb{P}=\{\textbf{P}_1,\textbf{P}_2,\textbf{P}_3...\textbf{P}_n\}$} denote the patches in $\textbf{V}$. For each $\textbf{P}_i$ a separate residual network $\bm{\phi}^R_i$ is learned to map $\textbf{P}_i$ to a face representation $\textbf{x}_i$, where $\textbf{x}_i = \bm{\phi}^R_i(\textbf{P}_i)$. Correspondingly, $\textbf{O}$ is also divided into equal sized patches as $\textbf{P}_i$, denoted by $\textbf{M}_i$. A binary value $m_i$ is used to summarize the occlusion status of $\textbf{M}_i$. In DPRFS, if the number of visible pixel in $\textbf{M}_i$ is smaller than a threshold $\epsilon$, the whole patch is regarded as occluded so that $m_i=0$. The final output of DPRFS is a concatenated face representation $\textbf{x} = [\textbf{x}_1, \textbf{x}_2,...,\textbf{x}_n]$ along with the corresponding occlusion vector $\textbf{m} = [m_1, m_2,..,m_n]$.

\subsection{Occlusion-guided compact template learning} 
\begin{figure}[!t]
		\begin{center}
			\begin{tabular}{cc}
				\includegraphics[width=3.7cm, height=2.8cm]{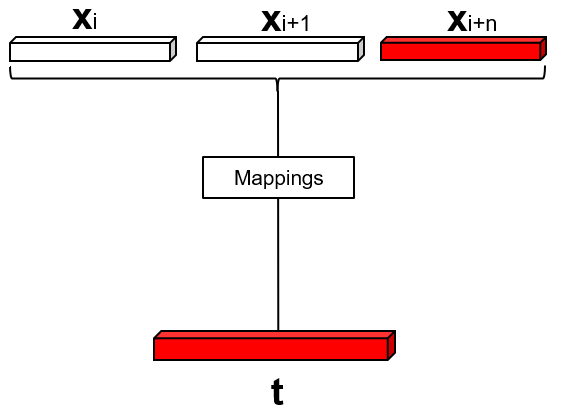} &
				\includegraphics[width=3.7cm, height=2.8cm]{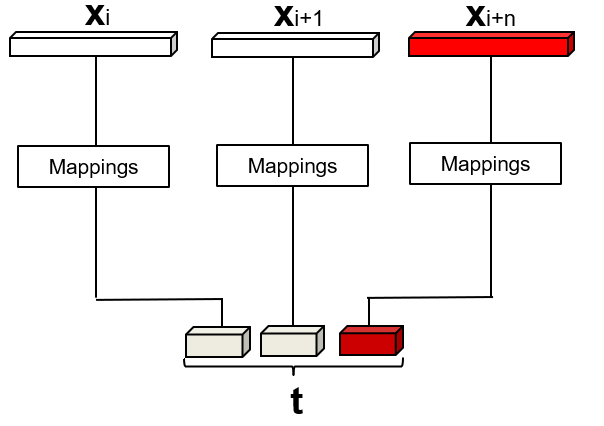} 
			\end{tabular}
		\end{center}
			\vspace{-0.6cm}
		\caption{Network architectures widely employed to derive a compact template from concatenated face representations. Templates $\textbf{t}$ generated from either architecture are sensitive to non-discriminative facial representations $\textbf{x}_{i+n}$ generated from occluded facial patches. (L) Network architecture employed in \cite{Sun_2014_16640, Sun_2014_17724, Sun_2015_16639,Chellapa}. (R) Network architecture employed in \cite{Liu_2015_16644}. }
		\label{Network}
		\vspace{-0.6cm}
	\end{figure}

To obtain a compact template $\textbf{t}$ from the face representations $\textbf{x}$, where the length of $\textbf{t}$ is at least a magnitude smaller than $\textbf{x}$, a mapping $\bm{\phi}_C$ from the face representations to the low dimensional template space has to be created so that $\textbf{t} = \bm{\phi}^C(\textbf{x})$. Some of the previous solutions map $\textbf{x}$ to a low dimensional space with fully connected layers \cite{TPN}, PCA \cite{Sun_2014_16640, Sun_2014_17724, Sun_2015_16639} or metric learning \cite{Chellapa}. The architecture of these works is summarized in Fig. \ref{Network}(L). Another solution \cite{Liu_2015_16644} generates a compact representation $\textbf{t}_i$ for each individual $\textbf{x}_i$ with individual fully connected layers $\textbf{t}_i = \bm{\phi}_i^C(\textbf{x}_i)$ and then concatenate them as $\textbf{t} = [\textbf{t}_1,\textbf{t}_2...,\textbf{t}_n]$. This architecture is depicted in Fig. \ref{Network}(R). Assuming a facial representation $\textbf{x}_{i+n}$ is generated from an occluded facial patch which contains identity-irrelevant information, the corresponding template $\textbf{t}$ will be contaminated by the identity-irrelevant information and result in a less discriminative template. Our experiments indicate that the compact templates generated from these architectures are suboptimal for face recognition under large head pose variations. 

\begin{figure}[!t]
	\begin{center}
		\begin{tabular}{c}
			\includegraphics[width=6.5cm, height=3.5cm]{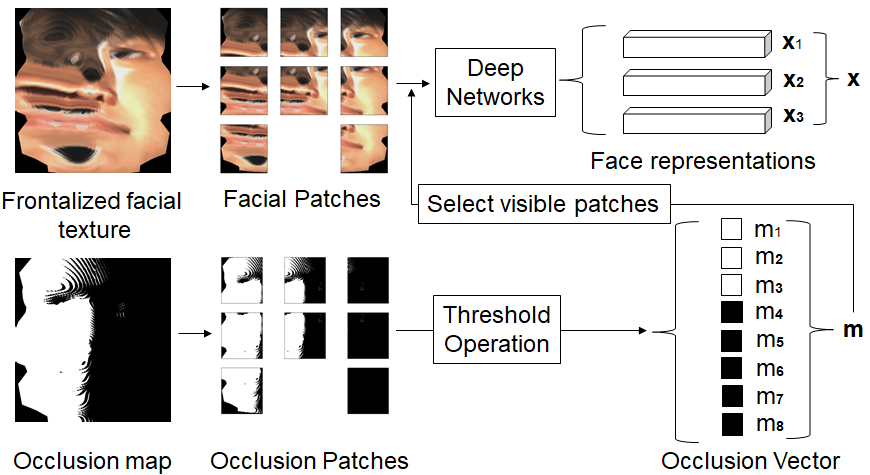} \\
		\end{tabular}
	\end{center}
	
	\caption{The DPRFS algorithm for encoding the input image into a concatenated face representation $\textbf{x}$ and occlusion vector $\textbf{m}$.  The white regions in the occlusion map indicate occluded regions, which correspond to $m_4,m_5,m_6,m_7$, and $m_8$ in the occlusion vector.}
	\vspace{-0.6cm}
	\label{DPRFS}
	
\end{figure}

	\textbf{The role of the occlusion mask:}  When a facial patch is occluded, it should have zero contribution to the final compact template. One straightforward solution to enforce this constraint is to directly convert the corresponding entries in the compact template space to zero. However, this will increase the intra-class variation when matching a frontal face to a profile face, because the distance between a zero-valued face representation generated from a profile face and a real-valued face representation from a frontal face does not reflect the actual similarity of the two faces. A possible solution proposed in DPRFS is to use  $\textbf{m}$ in template matching. The distance $s$ between facial image $a$ and $b$ in DPRFS can be computed as:
	\begin{align*}
	s= \dfrac{1}{\sum\limits({\textbf{m}^a\land \textbf{m}^b})}\sum\limits_i\Psi_{i}(\textbf{x}_i^a,\textbf{x}_i^b)({m_i^a\land m_i^b})
	\numberthis
	\end{align*}
	where $\Psi_{i}(\textbf{x}_i^a,\textbf{x}_i^b)$ represents the cosine distance between two face representations $\textbf{x}_i^a$ and $\textbf{x}_i^b$. The score $s$ records the average cosine distance of all face representations computed from visible patches. The symbol $\land$ represents the logical conjunction operation.

	Despite its robustness to facial occlusion, this method has two limitations: (i) the face retrieval speed is significantly reduced by computing $n$ times the patch-wise distance (once for every pair of patches:  $\Psi_{i}(\textbf{x}_i^a,\textbf{x}_i^b)({m_i^a\land m_i^b})$). (ii) the templates generated from an image-set are hard to be matched with $\textbf{m}$ because every individual sample in an image set may have a different $\textbf{m}$. To improve the matching speed and at the same time preserve the discriminative capability of the template in matching two image-sets, it is important to incorporate the occlusion mask in learning the compact template as $\textbf{t} = \bm{\phi}^C(\textbf{x},\textbf{m})$, and the distance between image $a$ and $b$ can be computed simply by $s = \Psi(\textbf{t}^a, \textbf{t}^b)$.

	\textbf{Co-occurrence information in template learning:} To learn a robust and discriminative feature representation for object recognition, encoding of the spatial co-occurrence among features increases the discriminative power of the final feature representation \cite{PRICoLBP}. The term ``contextual" is used in \cite{CALBFL, context} to describe the co-occurrence pattern between neighboring features and the authors claim that the contextual information encoded in facial patches enhances the robustness and stableness of face representation. In this paper, we present an approach that encodes the discriminative co-occurrence information in the space of $\textbf{t}$ using the face representations from visible patches. 

\begin{figure}[!t]
		\begin{center}
			\begin{tabular}{c}
				\includegraphics[width=6cm, height=3.3cm]{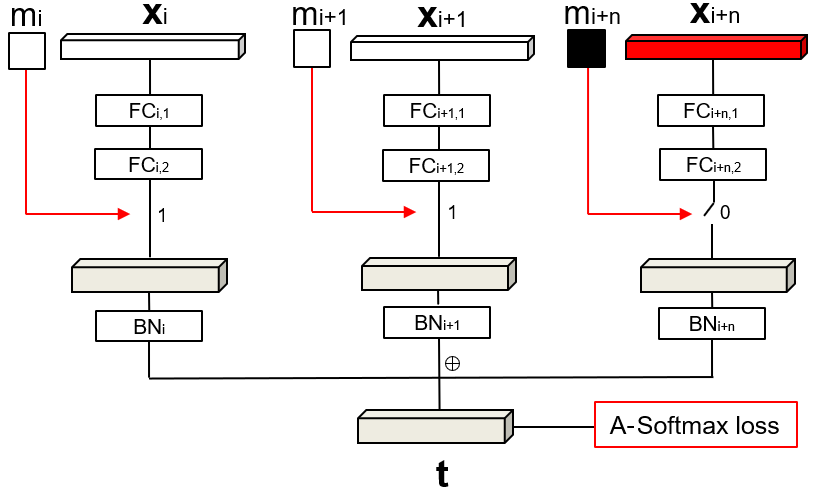}  \\
			\end{tabular}
		\end{center}
			\vspace{-0.6cm}
		\caption{Depiction of the network architecture employed in OGCTL. The occlusion mask controls the information flow from each face representation. }
		\vspace{-0.5cm}
		\label{OGCTL}
	\end{figure}

	It is assumed that the $\textbf{x}_i$ can be mapped into a co-occurrence space using non-linear projection $\bm{\phi}^D_i$ with parameters $\bm{\theta}^D_i$ so that the face representations can be added together and create the $\textbf{t}$. The reason we employ the addition operation to generate the co-occurrence information is because such an operation will use relatively smaller space than a concatenation operation but is also able to encode the co-occurence information from all the facial representations. To incorporate the information from the occlusion mask, $m_i$ is extended to be a vector that has the same length as $\textbf{x}$, denoted as $\textbf{m}_i$. Element values in $\textbf{m}_i$ simply repeat the value of $m_i$. Here are the equations to compute $\textbf{t}$:
	\begin{align*}
		& 	\hat{\textbf{t}}_i= \bm{\phi}^N_i({\textbf{m}_i\bm{\phi}^D_i(\textbf{x}_i; \bm{\theta}^D_i)};\bm{\theta}^N_i) \numberthis \\
			& \textbf{t} = \sum\limits_i\hat{\textbf{t}}_i \numberthis \>\>\>\>\>\>\>\>\>\>\>\>\>\>\>\>\>\>\>\>\>\>\>\>\>\>\>\>\>\>\>\>\>\>\>\>\>\> \CommaPunct
	\end{align*}
	where the non-linear mapping $\bm{\phi}^D_i$ transforms a face representation to a compact space. The mapping $\bm{\phi}^N_i$ indicates a normalization function to improve the convergence during optimization. The parameters optimized in training are  $(\bm{\theta}^D_i, \bm{\theta}^N_i)$. By using occlusion masks in template construction, the impact of an occluded patch to the output template would be a constant vector. The constant vector is invariant to the non-discriminative information encoded in the occluded patch since the information flow from the occluded patch is completely blocked in OGCTL. As a result, the identity irrelevant variations caused by the occluded patches are controlled and upper bounded. In the implementation, a two-layer fully connected network is employed with a bottleneck hidden layer and PreLU \cite{PRelu} activation function to be the non-linear mapping $\bm{\phi}^D_i$. Batch normalization \cite{Ioffe_2015_17038} is employed to be $\bm{\phi}^N_i$. The network architecture is depicted in Fig. \ref{OGCTL}.

	\textbf{A local to global mapping controlled by occlusion mask:} In \mbox{Eq. 2}, the mapping $\bm{\phi}^D_i$ maps a local representation $\textbf{x}_i$ to a global representation $\hat{\textbf{t}}_i$ so that it can be aggregated with other global representations. Thus, $\hat{\textbf{t}}_i$ can be treated as a global guess towards $\textbf{t}$ that is generated from local representation $\textbf{x}_i$, and each $\textbf{x}_i$ creates an independent guess towards $\textbf{t}$. When the $i^{th}$ patch is occluded, $\textbf{m}_i$ will mask out the invalid information coming from $\bm{\phi}^D_i(\textbf{x}_i)$ and block its impact to $\textbf{t}$.

	\textbf{Magnitude invariant loss function:} Because $\textbf{t}$ is a summation over  $\hat{\textbf{t}}_i$ that come from both visible and occluded patches, the number of occluded patches in Eq. 3 may affect the magnitude of $\textbf{t}$, denoted as $|\textbf{t}|$. Hence, a loss function that is insensitive to $|\textbf{t}|$ is required because we don't want the similarity of two samples to depend on the number of patches that are visible. In this work, we select the recently proposed angular softmax loss (A-Softmax) \cite{Liu_2017_17761} to learn the template embedding  $\textbf{t}$. Let $j$ denote the index of an image, and $y_j$ be the ground-truth label of the image. The loss on a training set $\{(\textbf{t}_j, y_j)|j\in[1,K]\}$ that contains $K$ samples is defined as:
	\begin{align*}
 &	L  =  - \dfrac{1}{K}\sum\limits_jlog(\dfrac{e^{f_1(\textbf{t}_j,\textbf{w}_{y_j},\omega)}}{e^{f_1(\textbf{t}_j,\textbf{w}_{y_j},\omega)}+\Sigma_{c \neq  y_j}e^{f_2(\textbf{t}_j,\textbf{w}_c)}}) \numberthis, \\
& f_1(\textbf{t}_j,\textbf{w}_{y_j},\omega)  = ||\textbf{t}_j||cos(\omega\theta_{j,y_j}) \numberthis, \\ 
& f_2(\textbf{t}_j,\textbf{w}_c) = ||\textbf{t}_j||cos(\theta_{j,c}), \numberthis
\end{align*}
where the symbol $\theta_{j,y_j}$ denotes the angle between template $\textbf{t}_j$ and the projection vector $\textbf{w}_{y_j}$ from the $y_j$ class. The projection vector $\textbf{w}$ is learned for each class in training, and the projection vectors compose a projection matrix $\textbf{W}$. The parameter $\omega$ controls the angular margin between classes. Because the decision boundary between class $y_j$ and class $c (c \neq  y_j)$ is only determined by $cos(\omega\theta_{j,y_j})- cos(\theta_{j,c})$, the decision boundary of A-softmax loss is not sensitive to the magnitude of $\textbf{t}$.

In summary, A-softmax loss enforces that templates $\textbf{t}$ originated from from the same class have small angular distance, while templates $\textbf{t}$ originated from different classes have large angular distance, regardless the number of visible patches that contribute to $\textbf{t}$. In testing, since the cosine distance is employed to compute the distance between templates, the magnitude of $\textbf{t}$ will have no impact in face matching. We name this algorithm Occlusion-Guided Compact Template Learning (OGCTL). The training portion is summarized in Alg. \ref{OGCTL_algorithm}.

\begin{algorithm}
	\KwIn{Training images $\mathbb{I}$ and label vector $\textbf{y}$}
	\KwOut{ $\{(\bm{\theta}^D_i, \bm{\theta}^N_i)|i\in[1,n]\}$ and $\textbf{W}$}
		\For  {each image $\textbf{I}$ in $\mathbb{I}$} {
	Crop facial patches $\mathbb{P}$ from $\textbf{I}$ and estimate their occlusion vector $\textbf{m}$\;
	Extract $\textbf{x}$ from $\mathbb{P}$ based on  $\{\bm{\phi}^R_i|i\in[1,n]\}$ \;
	Save $\textbf{m}$ and $\textbf{x}$ into the corresponding set $\mathbb{M}$ and $\mathbb{X}$  \;}
	Optimize $\{\bm{\theta}^D_i, \bm{\theta}^N_i\}$ in $\{\bm{\phi}^N_i$,$\bm{\phi}^D_i\}$ with $\{\mathbb{M}, \mathbb{X}, \textbf{y}, $L$\}$;
	\caption{Train the neural network in OGCTL}
	\label{OGCTL_algorithm}
	
\end{algorithm}

		\vspace{-0.5cm}
\section{Experiments}
The proposed algorithms are evaluated on the UHDB31 \cite{Le_2017_17704} and IJB-C \cite{IJBC} databases. We selected these databases because they contain large head pose variations and are still considered as challenging face recognition databases. In DPRFS, the threshold of occlusion $\epsilon$  is set to be 0.7. The eight ($n$=8) face representations generated from DPRFS are employed as the input of OGCTL. The number of hidden units in the two fully connected layers $\bm{\phi}^D_i$ is set to be 64. The $\omega$ in loss functions of OGCTL is set to be four as mentioned in the A-softmax loss paper \cite{Liu_2017_17761}.  ADAM optimizer \cite{Kingma_2015_17298} is used for optimizing the parameters in the networks, it takes 30 epochs to train the network in OGCTL. 
	
\textbf{UHDB31 database \cite{Le_2017_17704}:} This database contains 24,255 images from 77 subjects, captured under three illumination and 21 pose variations. The yaw angle variation is up to \ang{90} and the pitch angle variation is up to \ang{30}. In experiments, algorithms are evaluated under all three illuminations (I03, I01, and I05) with image resolution 128 $\times$ 153. Abbreviation ``I01" is used to denote the data partition with illumination 01. Partition ``I03" is employed for evaluating the baseline models; the best-performing models are selected based on this partition to show the results and are used for the other experiments. The 1{:}N face identification protocol proposed by Wu \etal \cite{Wu_2016_16536} is employed for evaluating algorithms on this database. Under this protocol, 77 frontal view images from each subject are selected to be the gallery, other images under different head pose variations are selected to be the probe images.

\textbf{IJB-C database \cite{IJBC}:} This recently released IARPA Janus Benchmarks database is a super-set of IJB-A \cite{Klare_2015_17419} and IJB-B \cite{IJBB} database where all images and videos were captured from unconstrained environments.  It contains 3,531 subjects with 31.3 \textit{K} still images and 117.5 \textit{K} frames from 11,779 videos. Models are evaluated on the standard 1:1 verification protocol, where a facial template is generated from a variable number of face images and video frames from different sources. In total, there are 23,124 templates with 19,557 genuine matches and 15.6 \textit{M} impostor matches.

\subsection{Template size in face identification}

The objective of the first experiment is to test OGCTL's performance under different sizes of output template. We change the embedding layers of deep neural networks so that the rank-1 identification performance of each method can be reported with respect to a various sizes of output templates. Multiple template generation algorithms are also compared in this experiment, which are illustrated below.

\begin{figure*}[tp!]
	\begin{center}
		\begin{tabular}{c}
	     	\includegraphics[width=14.cm,height=4.5cm]{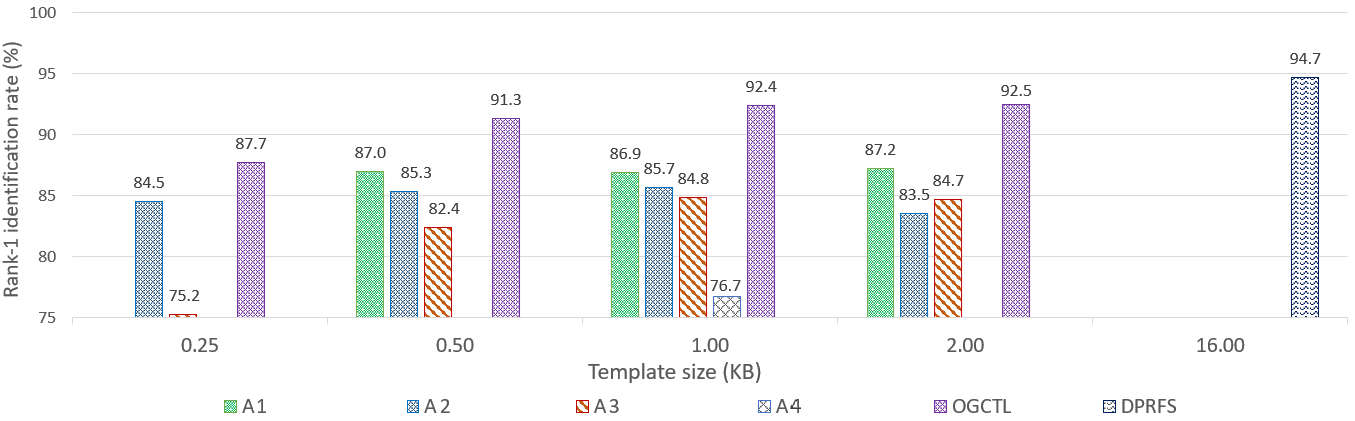} \\
			\includegraphics[width=14.1cm,height=4.5cm]{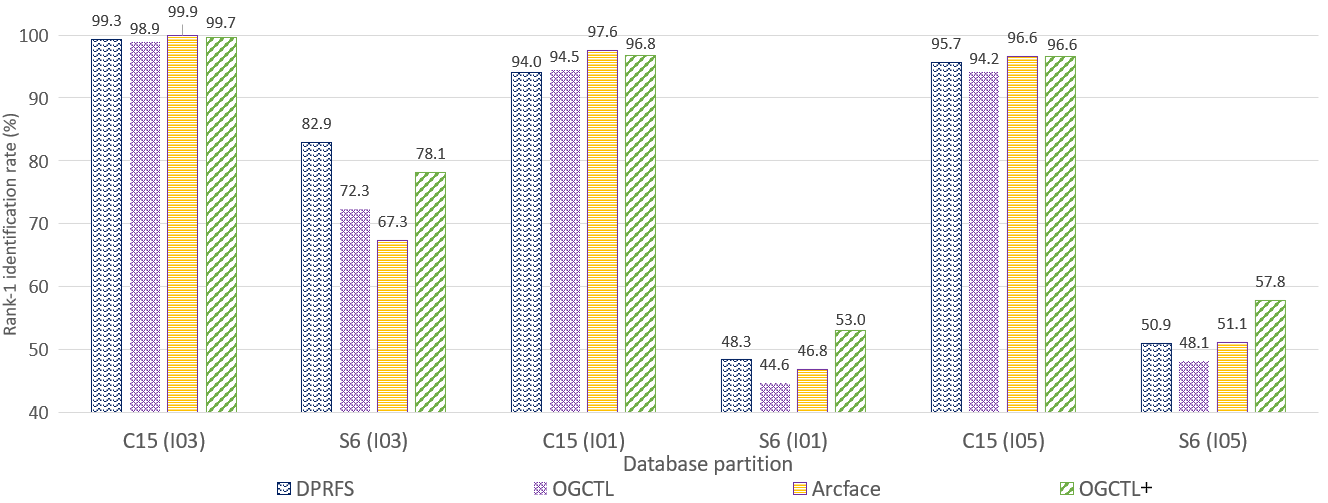} 
		\end{tabular}
	\end{center}
	\vspace{-0.7cm}
	\caption{Baselines' performance with respect to: (T) different data partitions on UHDB31 database, (B) different template sizes on UHDB31 database. Note that in ``S6'' partitions, all the faces are partially occluded.}
	\label{f3}
	\vspace{-0.4cm}
\end{figure*}

\noindent
\textbf{DPRFS}: This baseline is the original implementation of DPRFS \cite{Xu_2017_17643}. It employs eight ResNet-24 \cite{He_2016_17189} for face recognition.  As a reference, the template size of DPRFS is fixed to be the same as in the original implementations \cite{Xu_2017_17643}, which is 16 \textit{KB}. \\
\textbf{A1}: This baseline employs one ResNet-24 (has the same number of layers as DPRFS) to directly extract the facial template from an aligned image as \cite{Liu_2017_17761}. A-Softmax loss (AS) is employed to train the network. \\
\textbf{A2}: This baseline use the same network as A1 but uses softmax as the loss function. \\
\textbf{A3}: This baseline first converts face representation of DPRFS to a compact face representation with two fully connected layers (with PreLU activation function), and then concatenates them to be the final template, as depicted in Fig. \ref{Network}(R).  It resembles the network architecture used in \cite{Liu_2015_16644} but uses AS in learning.  \\
\textbf{A4}: This baseline employs two fully-connected layers to transform the concatenated face representation generated by DPRFS into a compact template, as depicted in Fig. \ref{Network}(L). It resembles the architecture used in \cite{TPN}.   

In Fig. \ref{f3}(T), it can be observed that OGCTL significantly outperforms architectures A1-A4 in all template size variations, which demonstrates the advantages of OGCTL over other baseline models. Its performance saturates after the template size reaches 1 \textit{KB} (128 dimensions floating point). Since there is only slight performance difference between the 0.5 \textit{KB} and 1 \textit{KB} template, in the following experiments, the 0.5 \textit{KB} template generated by OGCTL is employed as the standard output of OGCTL. 

\textbf{Template matching speed:}
With a Intel 6700K CPU, the template matching speed of DPRFS is below 50 \textit{K} templates per second, the actual matching speed depends on the occlusion pattern of the probe image. The matching speed of OGCTL is more than 1 \textit{M} templates per second, it is independent from the occlusion pattern of the probe image.

\subsection{Test under pose and illumination variations}
The objective of the second experiment is to evaluate algorithms' performance under pose and illumination variations. We employ all three partitions of UHDB31 database, with the notation ``I03'', ``I01'', and ``I05''. Each partition has its own illumination condition. For each partition, the database is further divided into two sub-partitions: C15 and S6. In C15, the yaw angles of facial images are equal or below \ang{60}. C15 contains 15 pose variations. In S6, the yaw angles of facial images are equal to \ang{90}. Thus, S6 contains six pose variations. The average rank-1 identification rate of every pose in each sub-partition is reported in Fig. \ref{f3}(B). It is observed that the merits of DPRFS over OGCTL manifest themselves in the normal illumination condition (I03). In addition, we report the performance of Arcface \cite{ArcFace} in Fig. \ref{f3}(B), which uses a single ResNet-50 network to generate \mbox{2 \textit{KB}} template. It can be observed that Arcface performs better than DPRFS in C15 partition but outperforms by DPRFS in S9 partition (except I05). This experiment demonstrated that the ensemble network-based model (DPRFS) holds its merit when the pose variation is large. 

To boost the performance of OGCTL further under challenging illumination and pose variations, we merge the template of DPRFS and Arcface but preserving the output template size. Specifically, every face representation $\textbf{x}_i$ of DPRFS is concatenated with the template of Arcface and then feed into the network of OGCTL. This method is named OGCTL+. From Fig \ref{f3}(B), it is observed that OGCTL+ performs significantly better than OGCTL under large head pose variations with challenging illumination conditions. This experimental result demonstrated that the OGCTL is able to take advantage of the templates from both ensemble network model and a well-trained single network to generate a compact template for unconstrained face recognition.

\subsection{Image-set based face verification}
The objective of the third experiment is to further examine the performance of OGCTL and its generalization in compact template construction. In this experiment, we evaluated the algorithms presented under the challenging IJB-C database.
The widely adopted average pooling operation \cite{Wang_2017_17758,Xu_2017_17643,IJBC} is used to merge the templates from different persons into a single template for verification. The result is shown in Table \ref{ROC}. Baselines were cited from the original IJB-C benchmark \cite{IJBC}. It can be observed from Table \ref{ROC} that with a relatively small template size and a small training database, OGCTL is able to achieve improved performance than baselines that trained with a larger database with increased template size. Compared to DPRFS, because the occlusion mask is encoded in the template construction process of OGCTL, the average pooling operation only work on the discriminative co-occurrence information encoded in the visible patches, which results in a more discriminative template. Compared to OGCTL, the template of OGCTL+ are constructed based on both DPRFS and Arcface. It outperforms Arcface in terms of AUC, achieves new state-of-the-art performance, but with only 0.5 \textit{KB} template size. 

\subsection{Ablation study}
The objective of the fourth experiment is to analyze the usefulness of each proposed module in OGCTL and examine the unique properties of the template generated by OGCTL. The experiment is conducted on the UHDB31 database ``I03'' partition. We select the \mbox{Pose \#20} to report the face identification performance. The probe images in \mbox{Pose \#20} contain profile faces whose yaw variation is equal to \ang{90}. Gallery images are all frontal faces. The template generated by OGCTL is compared to DPRFS and A3. The A3 is selected because it can be regarded as an learning algorithm without using the occlusion mask. We also compare an application of OGCTL without A-softmax loss but using softmax loss in learning. This baseline is denoted as OGCTL-S. The results are depicted in Table \ref{Ablation}.

\begin{table*}[tp!]
	\centering
	\caption{TAR (True Accept Rate) against the FAR (False positive rate) of different methods on IJB-C database. The number of images used to train each model is presented under the ``scale". The AUC represents the Area Under the ROC. The result marked with $\dagger$ are read from the ROC curve in the IJB-C benchmark \cite{IJBC}. An unavailable number is marked as ``-''.}
	
	\scalebox{0.68}{
		\begin{tabular}{|c|c|c|c|c|c|c|c|c|c|c|}
	        \hline
			\textbf{Method} & \textbf{Input} & \textbf{Template}  & \textbf{Dataset} & \textbf{Scale} & \textbf{FAR}=$10^{-5}$ & \textbf{FAR}=$10^{-4}$ & \textbf{FAR}=$10^{-3}$ & \textbf{FAR}=$10^{-2}$ & \textbf{FAR}=$10^{-1}$ & \textbf{AUC} \\  & & \textbf{Size} & & & & & & & & \\ \hline
			COTS-1 {\cite{IJBC}]}$\dagger$ & Image & -         & -    & -            & 0.090    & 0.160    & 0.320    & 0.620    & 0.800 & -  \\ \hline
			FaceNet {\cite{IJBC}} $\dagger$ & Image & \textbf{0.5 \textit{KB}}     & -       & -        & 0.330    & 0.490    & 0.660    & 0.820    & 0.920 & -   \\ \hline
			VGG-CNN {\cite{IJBC}} $\dagger$ & Image & 16 \textit{KB}      & VGGFace & 2.6M  & 0.430    & 0.600    & 0.750    & 0.860    & 0.950 & -  \\ \hline
			Cao \etal \cite{VGGFace2}  & Image & 2 \textit{KB}     &  VGG2 & 3.3M                & 0.747    & 0.840    & 0.910    & 0.960    & 0.987 &   - \\ \hline
			Arcface \cite{ArcFace}  & Image & 2 \textit{KB}      & VGGFace & 2.6M  & \textbf{0.895}    & \textbf{0.932}    &  \textbf{0.957}    & 0.973    & 0.985 & 0.993  \\ \hline\hline
			
			DPRFS {\cite{Xu_2017_17643}} & Frontalized Image  & 16 \textit{KB}      & Webface & 0.5M  & 0.310    & 0.461    & 0.638    & 0.807    & 0.939  & 0.976 \\ \hline
			OGCTL     & Face representation     & \textbf{0.5 \textit{KB}}     & Webface & 0.5M  & 0.608    & 0.737    & 0.839    & 0.918    & 0.975   & 0.989 \\ \hline
			OGCTL+  & Face representation      & \textbf{0.5 \textit{KB}}       & Webface + MS-Celeb + VGG2 & 14M & 0.879    & 0.923    & 0.952    &  \textbf{0.975}    &  \textbf{0.988}  &  \textbf{0.995}  \\ \hline
		\end{tabular}
		\label{ROC}
		
	}
\end{table*}
The methods are compared under two experimental settings. In the first setting, the facial templates of both gallery and probe are constructed from the three visible patches as probe. In the second setting, the templates of gallery are constructed from all the eight visible patches but the templates of probe are constructed from three visible patches. The comparison between A3 and OGCTL highlights that the occlusion-guided element adding operation in OGCTL significantly improves the discriminative capability of the face template. The comparison between OGCTL-AS and OGCTL highlights that the A-softmax loss is essential in preserving the cosine distance after element adding operation. The comparison between DPRFS and OGCTL demonstrated that OGCTL is able to generate a template from different number of facial patches, taking into account the co-occurrence information between patches but preserving the compactness of the output template. 

\begin{table}[tp!]
	\centering
	\caption{Performance in profile face identification. The template size marked with $\dagger$ is determined by the number of visible patches (depicted in Fig. \ref{patches}) used for matching.}
	
	\scalebox{0.7}{
		\begin{tabular}{|c|c|c|c|c|}
			\hline
	        \textbf{Method} & \textbf{Gallery Patch} & \textbf{Probe Patch} & \textbf{Template} & \textbf{Identification} \\ & \textbf{Number} & \textbf{Number} & \textbf{Size} & \textbf{Accuracy} \\ \hline
			DPRFS  & 3   & 3              & 6.00 \textit{KB} $\dagger$    & 0.961     \\ \hline
			A3 & 3    &   3  & 0.19 \textit{KB}  $\dagger$   & 0.779    \\ \hline
			OGCTL-S & 3    & 3               & 0.50 \textit{KB}    & 0.922    \\ \hline
			OGCTL & 3   & 3  & 0.50 \textit{KB}    & 0.935    \\ \hline\hline
			
			OGCTL-S & 8     & 3   & 0.50 \textit{KB}    & 0.883    \\ \hline
			OGCTL & 8       & 3   & 0.50 \textit{KB}    & \textbf{0.974}   \\ \hline
		\end{tabular}
		
		\label{Ablation}
		
	}
\end{table}

\vspace{-0.2cm}
\section{Conclusion}
We presented OGCTL - an algorithm to construct a compact facial template based on the output of ensemble deep network models for efficient storage and fast face retrieval. The compact template integrates the face representations of visible facial patches through the proposed occlusion-guided compact template learning approach. The proposed algorithm provides a solution to generate reduced sized template with similar or better face recognition performance from ensemble networks.
\begin{figure}[!t]
	\begin{center}
		\begin{tabular}{c}
			\includegraphics[width=4.2cm, height=1.5cm]{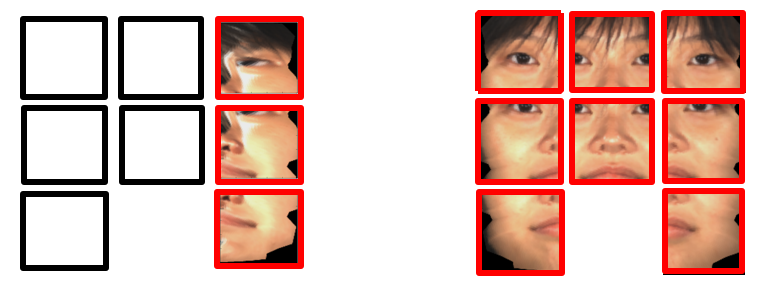} \\
		\end{tabular}
	\end{center}
	\vspace{-0.6cm}
	\caption{Facial template is able to be constructed from different number of facial patches. (L) Visible patches in profile face, and (R) Visible patches in frontal face. }
	\vspace{-0.6cm}
	\label{patches}
\end{figure}

	\vspace{-0.2cm}
\section{Acknowledgment}
This material is based upon work supported by the U.S. Department of Homeland Security under Grant Award Number 2017-ST-BTI-0001-0201. This grant is awarded to the Borders, Trade, and Immigration (BTI) Institute: A DHS Center of Excellence led by the University of Houston, and includes support for the project	``EDGE" awarded to the University of Houston. The views and conclusions contained in this document are those of the authors and should not be interpreted as necessarily representing the official policies, either expressed or implied, of the U.S. Department of Homeland Security.

{\small
\bibliographystyle{ieee}
\bibliography{ab,Mendeley_room314_Shared,lazyYuhang}
}

\end{document}